\documentclass[letterpaper, 10 pt, conference]{ieeeconf}  

\IEEEoverridecommandlockouts                              

\usepackage{xargs}
\usepackage{float}
\usepackage{wrapfig}
\usepackage{amsmath,amsfonts}
\usepackage{algorithm}
\usepackage{array}
\usepackage{textcomp}
\usepackage{stfloats}
\usepackage{url}
\usepackage{verbatim}
\usepackage{graphicx}
\usepackage{cite}
\usepackage{xcolor}

\usepackage{graphicx}
\usepackage{caption}
\usepackage{subcaption}
\usepackage{svg}
\usepackage{algorithm, setspace}
\usepackage[noend]{algpseudocode}
\usepackage[%
    colorlinks=true,
    pdfborder={0 0 0},
    linkcolor=red
]{hyperref}

\algnewcommand\algorithmicforeach{\textbf{for each}}
\algdef{S}[FOR]{ForEach}[1]{\algorithmicforeach\ #1\ \algorithmicdo}

\usepackage{graphics} 
\usepackage{epsfig} 
\usepackage{times} 
\usepackage{amsmath} 
\usepackage{amssymb}  
\usepackage{textcomp, gensymb}
\usepackage{url}
\usepackage{graphicx}
\usepackage{color}
\usepackage{xcolor}
\usepackage{soul}
\usepackage{mathtools}
\usepackage{algorithm, setspace}
\usepackage[noend]{algpseudocode}
\usepackage{ccicons}
\usepackage{comment}
\usepackage{bbm}
\usepackage{multirow}
\usepackage[compact]{titlesec}
\usepackage{booktabs} 
\renewcommand{\baselinestretch}{0.99}

\newtheorem{theorem}{Theorem}
\newtheorem{lemma}{Lemma}
\newtheorem{corollary}{Corollary}

\newcommand{\vol}{\mathrm{vol}}

\usepackage[font={small}]{caption}

\newcommand{\qa}[1]{{\textcolor[HTML]{6491EA}{\textbf{#1}}}}

\title{\LARGE \bf
Safety Evaluation of Motion Plans Using Trajectory Predictors as Forward Reachable Set Estimators
}

\author{Kaustav Chakraborty$^{*1,2,3}$, Zeyuan Feng$^{*2}$, Sushant Veer$^{* 3}$, Apoorva Sharma$^{3}$, Wenhao Ding$^{3}$\\ Sever Topan$^{3}$, Boris Ivanovic$^{3}$, Marco Pavone$^{2, 3}$, Somil Bansal$^{2}$  
\thanks{*equal contribution, arranged in alphabetical order of last names. Corresponding author: kaustavc@usc.edu.}
\thanks{$^{1}$Authors are with the Department of Electrical Engineering,  University of Southern California, USA.}
\thanks{$^{2}$Authors are with the Department of Aeronautics and Astronautics, Stanford University, USA}
\thanks{$^{3}$Authors are with NVIDIA Research
        }
}

\begin{document}

\maketitle
\thispagestyle{empty}
\pagestyle{empty}

\begin{abstract}
%
The advent of end-to-end autonomy stacks---often lacking interpretable intermediate modules---has placed an increased burden on ensuring that the final output, i.e., the motion plan, is safe in order to validate the safety of the entire stack. 
This requires a safety monitor that is both complete (able to detect all unsafe plans) and sound (does not flag safe plans). 
In this work, we propose a principled safety monitor that leverages modern multi-modal trajectory predictors to approximate forward reachable sets (FRS) of surrounding agents. 
By formulating a convex program, we efficiently extract these data-driven FRSs directly from the predicted state distributions, conditioned on scene context such as lane topology and agent history.
To ensure completeness, we leverage conformal prediction to calibrate the FRS and guarantee coverage of ground-truth trajectories with high probability. 
To preserve soundness in out-of-distribution (OOD) scenarios or under predictor failure, we introduce a Bayesian filter that dynamically adjusts the FRS conservativeness based on the predictor’s observed performance.
We then assess the safety of the ego vehicle’s motion plan by checking for intersections with these calibrated FRSs, ensuring the plan remains collision-free under plausible future behaviors of others.
Extensive experiments on the nuScenes dataset show our approach significantly improves soundness while maintaining completeness, offering a practical and reliable safety monitor for learned autonomy stacks.

\end{abstract}

\section{INTRODUCTION}
Classical planning stacks come in various shapes and forms \cite{teng2023motion,schwarting2018planning}, but, importantly, they share the characteristic of optimizing an interpretable cost function which allows reasoning about the plan's safety during synthesis. 
With the ever-increasing adoption of learning-based planners and end-to-end robot stacks, safe-by-construction planning has become extremely challenging, if not outright impossible. 
Therefore, safety monitors for motion plans have grown in prominence for ensuring that these often uninterpretable learning-based plans are safe. 
Like any effective monitor, a safety monitor should satisfy two key properties: completeness (it must flag all unsafe plans) and soundness (it must not flag safe ones). 
Our objective in this paper is to develop a method that can improve on the soundness of reachability-based safety monitors without compromising on completeness. 

\begin{figure*}[t]
    \centering
    \begin{subfigure}[b]{0.32\textwidth}
        \includegraphics[width=\textwidth]{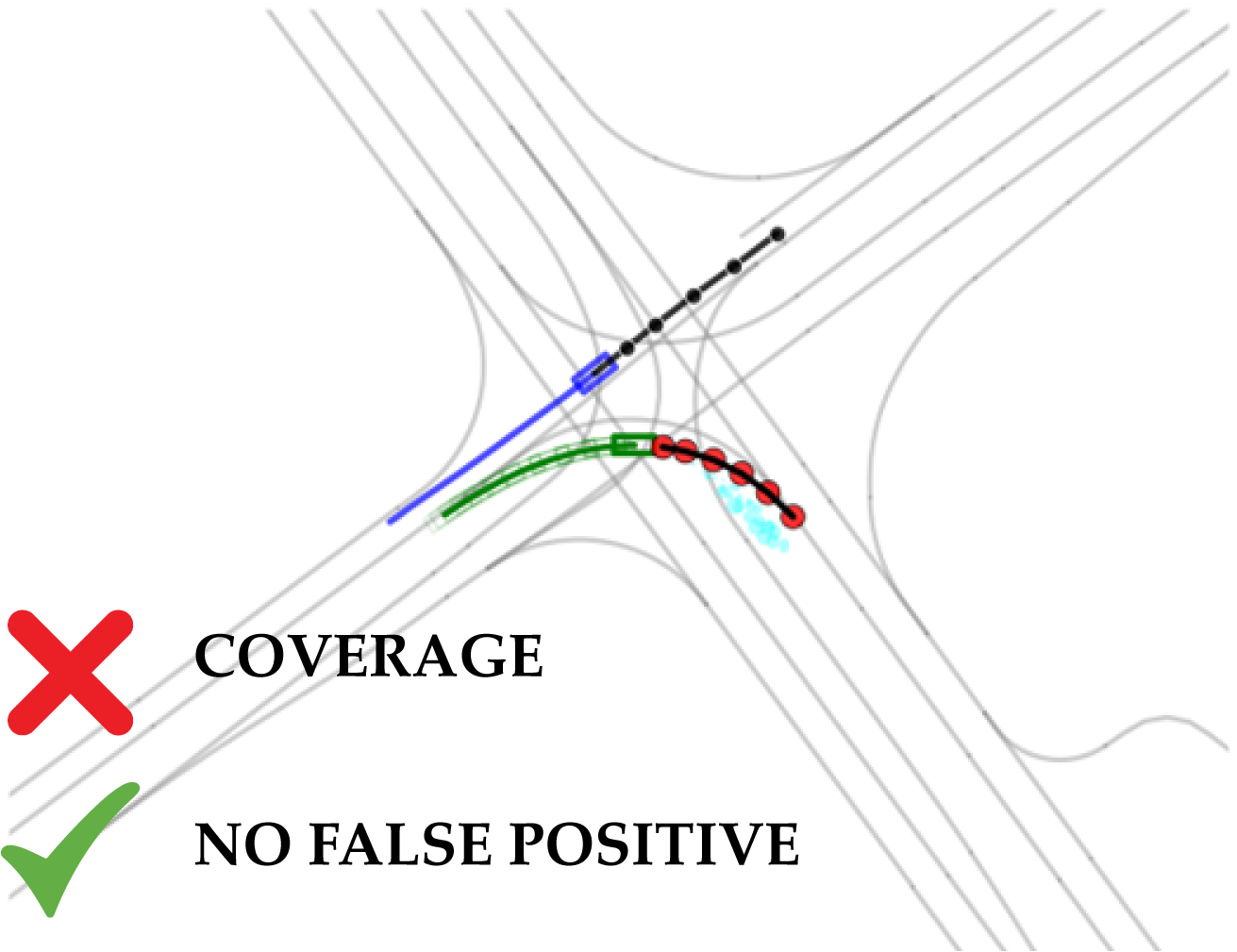}
        \caption{Baseline Predictor}
        \label{fig:subfig1}
    \end{subfigure}
    \hfill
    \begin{subfigure}[b]{0.32\textwidth}
        \includegraphics[width=\textwidth]{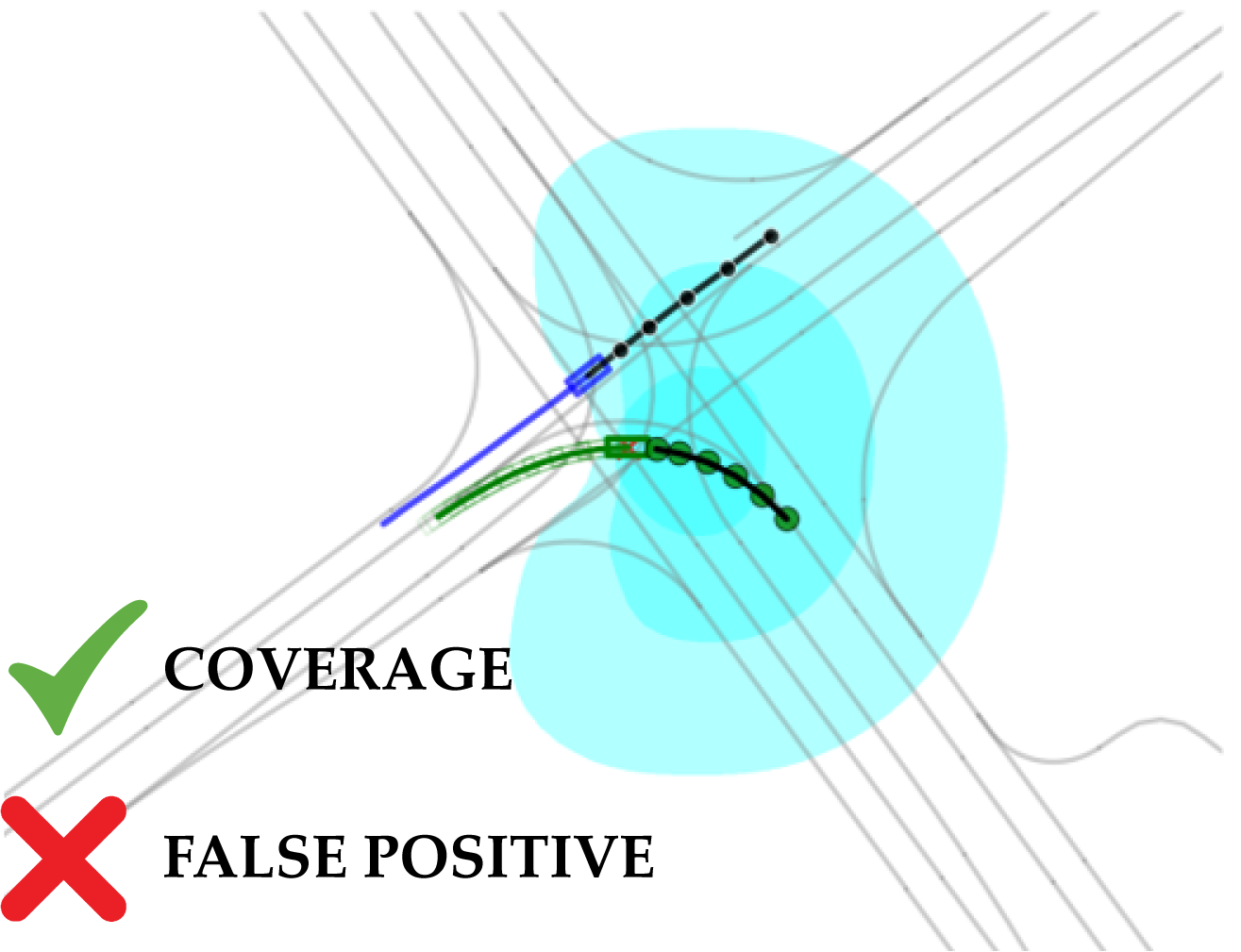}
        \caption{FRS}
        \label{fig:subfig2}
    \end{subfigure}
    \hfill
    \begin{subfigure}[b]{0.32\textwidth}
        \includegraphics[width=\textwidth]{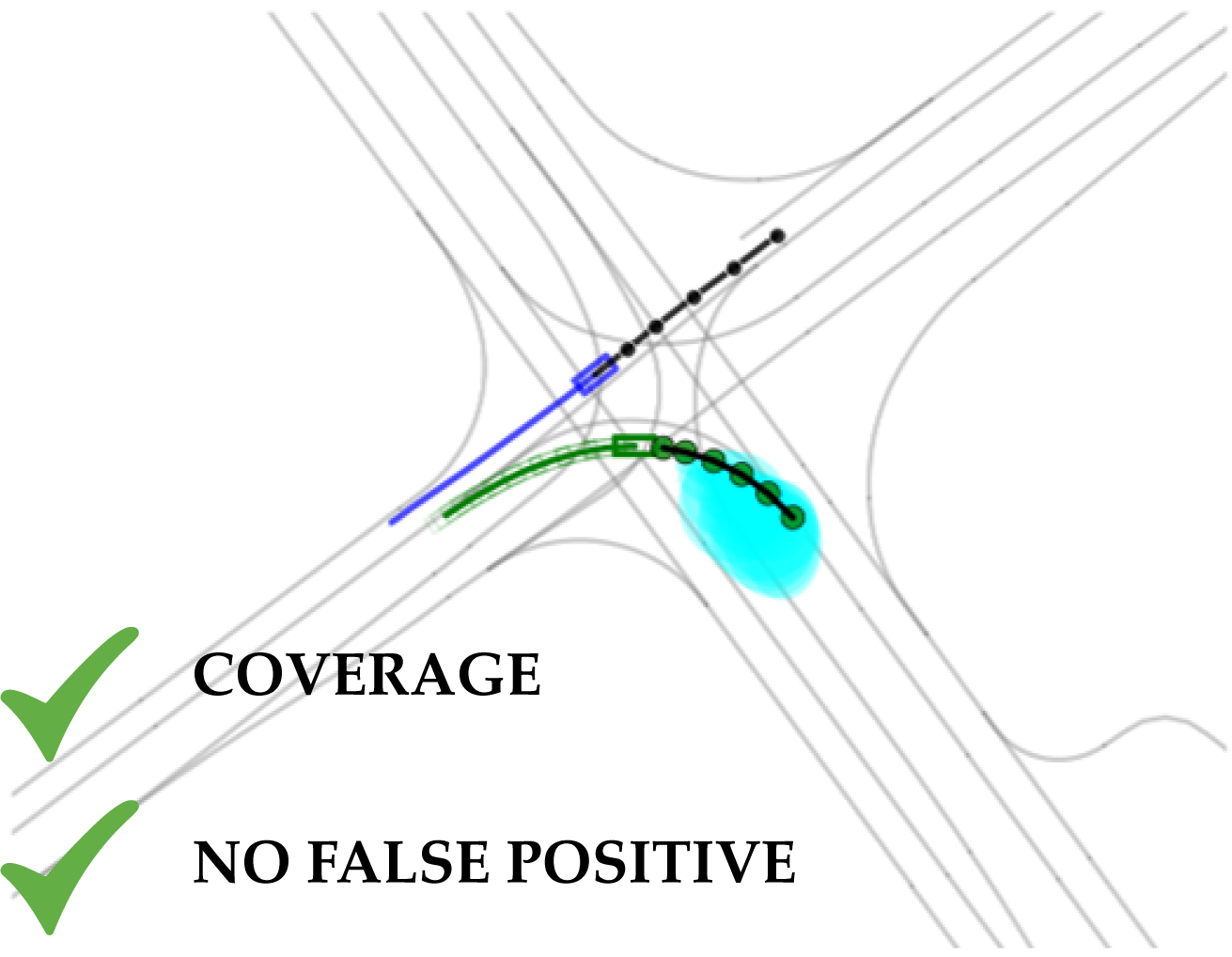}
        \caption{FORCE-OPT (ours)}
        \label{fig:subfig3}
    \end{subfigure}
    \caption{\small{In this intersection scenario from the nuScenes dataset, we evaluate the safety of the ego (blue) vehicle's planned trajectory (black) in the vicinity of a non-ego contender agent (green). Our objective is to verify that the ego’s planned path avoids any potential collisions with the contender by ensuring it remains outside all regions that the contender could possibly occupy in the future. We compare three methods of reachable set computation for safety evaluation of ego's motion plan.  \textbf{(a) Baseline Predictor:} This method produces very small, overconfident predictions (cyan ellipses) for the other agent's future positions. While it correctly avoids a false positive (FP) collision warning, it fails to cover the true future positions of the contender (the cyan ellipses do not cover the true ground truth states states shown with red dots), making it unreliable for safety---indeed, in the detailed results later in the paper (Section~\ref{sec:result}) this methods exhibits a high False Negative Rate (FNR). \textbf{(b) FRS (Forward Reachable Set):} This approach creates a large, conservative reachable set (the large cyan circle) for the other agent. It achieves excellent coverage, but its excessive size causes it to overlap with the ego vehicle's path, triggering a FP alarm. \textbf{(c) FORCE-OPT (ours): } Our method generates a more realistic and tighter reachable set (the smaller cyan area) for the other agent. It successfully achieves coverage by containing the agent's true future path (green dots) while being precise enough to avoid intersecting with the ego's trajectory, preventing a FP, correctly identifying the plan as safe. Overall, our method demonstrates a good balance of FPR and FNR as discussed in greater detail in Section~\ref{sec:result}.}}
    \label{fig:all_subfigures}
    \vspace{-5mm}
\end{figure*}

We achieve this by reinterpreting trajectory predictors - typically generative models trained to forecast future agent behavior conditioned on histories and scene context (e.g., lane graphs, traffic signals) - as \textit{data-driven forward reachable set (FRS) estimators}. 
These predictors, often instantiated as Gaussian Mixture Models (GMMs), implicitly learn a stochastic model of agent dynamics from logged driving data. Our \textit{key insight} is that we can extract a probabilistic FRS by solving an optimization problem that identifies the smallest-volume set that captures a desired amount of probability mass under the learned distribution.

Leveraging the fact that many modern trajectory predictors output GMMs on the future states \cite{salzmann2020trajectron++,girgis2021latent}, we formulate a convex relaxation of this problem that allows us to efficiently compute tight reachable sets that are significantly less conservative than traditional worst-case reachability approaches. However, these learned distributions are subject to modeling error and may fail to capture the true dynamics, especially under distribution shift.

To combat learning errors, we use conformal prediction (CP) \cite{angelopoulos2021gentle,shafer2008tutorial} to calibrate the FRS. 
Rather than inflating the predicted sets indiscriminately, CP allows us to scale the covariance of each GMM component just enough to ensure that the reachable set covers the ground-truth trajectory with a user-specified error rate.
Our overall algorithm to extract the FRS is called FORCE-OPT, which stands for \underline{FO}rward \underline{R}eachable sets from \underline{C}onformal \underline{E}stimation and convex \underline{OPT}imization. 

Although FORCE-OPT comes with a probabilistic guarantee on covering the ground truth when operating within the calibration distribution, it 
may still degrade in the presence of distribution shift. 
To handle this, we introduce a Bayesian filtering mechanism that monitors the consistency between predicted and observed agent behavior, adjusting the FRS conservativeness on the fly based on our confidence in the trajectory predictor.

This main contributions of the paper are: (i) We provide a rigorous formulation of the problem of estimating FRS from trajectory predictors. 
(ii) We introduce FORCE-OPT, an efficient algorithm that combines convex optimization and conformal prediction to compute calibrated, data-driven FRS from GMM-based predictors.
(iii) We incorporate a belief-based Bayesian filtering mechanism that adapts the reachable set dynamically to account for predictor reliability under distribution shift.
(iv) Through testing on nuScenes \cite{nuscenes}, real-world driving dataset, we demonstrate that FORCE-OPT achieves the best balance of false positive and false negative rates, outperforming both conservative baselines and uncalibrated learned methods.

\section{RELATED WORKS}

\textbf{Control-Theoretic Safety Evaluators.} 
Classical approaches for safety evaluation of motion plans have relied on control-theoretic tools such as formal verification \cite{luckcuck2019formal, mitra2024formal}, which provide mathematical guarantees that a system satisfies a predefined set of safety specifications. While rigorous, these methods often fall short when applied to modern robotic systems that encounter large uncertainties and operate in high-dimensional, stochastic environments. To handle this imminent uncertainty, reachability-based techniques—including those based on zonotopes \cite{althoff2011zonotope}, Hamilton-Jacobi formulations \cite{lygeros1999controllers,mitchell2005toolbox}, or sums-of-squares \cite{majumdar2017funnel}—aim to characterize the set of all future states a system can reach under bounded disturbances. However, their worst-case assumptions often lead to overly conservative safety bounds that limit practical usability. Furthermore, these methods typically struggle to incorporate rich sensory inputs (e.g., images, LiDAR) and contextual cues that modern autonomous systems depend on. High-dimensional observations are often abstracted through models \cite{fridovich2020confidence} or simplifying assumptions are made, reducing the granularity of the safety assessment. A common alternative to handling the difficult-to-model uncertainties in the real world is through data-driven methods. One such approach, tailored towards multi-agent settings, is to leverage learning-based trajectory predictors, as we will discuss next.

\textbf{Trajectory Prediction-based Safety Evaluators.} Motivated by the limitations of worst-case control-theoretic safety evaluators, there has been growing interest in safety evaluation frameworks built on top of learning-based trajectory predictors \cite{salzmann20, hwang2024emma,girgis2021latent}. 
Although trajectory predictors were devised to aid with planning \cite{karkus2023diffstack,chen2023interactive,casas2021mp3,cui2021lookout}, they have found widespread use in other applications, such as failure monitoring \cite{antonante2023task,chakraborty2024system}, mining interactive scenarios \cite{dinparastdjadid2023measuring,ding2025surprise}, and assessing planner safety \cite{nakamura2023online,lindemann2023safe,li2020prediction} which is of primary interest to us in this paper. Trajectory predictors are often used to identify reachable zones for other agents that the motion plan should stay out of, violation of which is considered to trigger a safety failure; see \cite{althoff2021set} for a survey. Methods in this category include those that use trajectory predictors to guide controllability bounds on other agents to be used in a classical reachability formulation \cite{nakamura2023online,li2020prediction,fridovich2020confidence} and those that directly extract a reachable zone for contender agents from the predictor \cite{paparusso2024zapp,lindemann2023safe,dixit2023adaptive,chen2021reactive}. In the latter category, approaches have explored fitting zonotopes to prediction outputs for estimating reachable sets \cite{paparusso2024zapp}, leveraged conformal prediction \cite{lindemann2023safe,dixit2023adaptive,chen2021reactive}, and solved a probabilistic optimization on the output distribution of the predictor \cite{driggs2017robust,devonport2020estimating}. However, many of these approaches require sampling \cite{tumu2024multi,xiang2024convex,dietrich2025data} which is computationally expensive, lack multi-modality \cite{lindemann2023safe}, or lack calibration of the predictors \cite{driggs2017robust,driggs2017integrating} resulting in learning-errors affecting the quality of the FRS. The approach presented in this paper is computationally efficient, accounts for multi-modality in predictions, calibrates the predictions to mitigate the impact of learning errors, and performs a belief-based FRS adaptation to impart some degree of OOD robustness. 



\section{Equivalence of Deterministic and Stochastic Notions of FRS}
We now present a rigorous mathematical framework to bridge the probabilistic and deterministic definitions of FRS. While the former is characteristic of contemporary trajectory forecasting literature, the latter is fundamental to control-theoretic safety analysis. 

Let $x_t\in \mathcal{X}\subseteq\mathbb{R}^{n_x}$ denote the state of an agent at discrete time $t\in\mathbb{Z}_+$. The agent’s control inputs are denoted by $u_t \in \mathcal{U}_t \subseteq \mathbb{R}^{n_u}$, while other disturbances arising from exogenous factors and uncertainties are captured by $w_t\in\mathcal{W}_t\subseteq\mathbb{R}^{n_w}$. 
The agent's dynamics evolve according to a discrete-time function $f: \mathbb{Z}_{+} \times \mathcal{X} \times \mathcal{U} \times \mathcal{W} \rightarrow \mathcal{X}$, which is assumed to be continuously differentiable in its arguments:
\begin{equation}\label{eq:dynamics}
    x_{t+1} = f(t,x_t, u_t, w_t).
\end{equation}
For notational brevity, we define $\hat{f}_{t,u_t,w_t}(x_t):=f(t,x_t, u_t, w_t)$. The \emph{forward-reachable set} (FRS) $F_t$ at time $t$, starting from an initial state $x_0\in\mathcal{X}$, is the set of all states that the agent can reach under a set of possible control actions and disturbances:
\begin{align}\label{eq:deterministic-FRS}
    F_t := & \{\hat{f}_{t-1, u_{t-1}, w_{t-1}}\circ \cdots \circ \hat{f}_{0,u_0, w_0}(x_0) \\ \nonumber
    &~|~u_i\in\mathcal{U}_i, w_i\in\mathcal{W}_i, i=0,\cdots,t-1\}
\end{align}
%
In practice, however, the precise control sets $\mathcal{U}_i$ and disturbance sets $\mathcal{W}_i$ are unknown and influenced by hard-to-model aspects such as driver intent, road geometry, and local context. 
Classical reachability methods often adopt worst-case assumptions over these sets, resulting in overly conservative FRSs. Such assumptions are unnecessarily pessimistic - for instance, it is unreasonable to expect that a stopped vehicle at a red light will suddenly accelerate through the intersection while the traffic light is still red.
A more realistic alternative is to infer the agent’s behavior from data. In what follows, we introduce a probabilistic formulation of the FRS that facilitates using data-driven trajectory predictors to estimate likely future states of an agent, thereby, reducing the conservatism of the FRS.

We now reformulate the FRS using a probabilistic lens, as inspired by prior work \cite{driggs2017robust}.
As we highlight later in this paper, this probabilistic viewpoint is readily compatible with modern multi-modal trajectory predictors.
Let $\vol$ represent the volume (Lebesgue measure) of a measurable set $\omega \in \Omega$, where $\Omega$ is the collection of all measurable subsets of $\mathcal{X}$. 
Let $\mu_t : \Omega \to [0, 1]$ represent a probability measure describing the distribution over the agent’s state at time $t$.
This measure arises as the push-forward of absolutely continuous probability distributions over the sequences of control and disturbance inputs, with support on $\mathcal{U}_i$ and $\mathcal{W}_i$, mapped through the composed dynamics in \eqref{eq:deterministic-FRS}.
The probabilistic FRS can then be defined through the following optimization problem:
\begin{equation}\label{eq:probabilistic-FRS}
    \omega_t^* := \underset{\{\omega\in\Omega: \mu_t(\omega) = 1\}}{\arg\inf} \vol(\omega).
\end{equation}
%
Intuitively, \eqref{eq:probabilistic-FRS} returns the smallest (i.e., minimal-volume) set that captures all the future states of the agent under the distribution $\mu_t$. Remarkably, this probabilistic formulation is equivalent to the classical deterministic definition in \eqref{eq:deterministic-FRS} almost everywhere (i.e., excluding a set of measure zero) as formalized in the following theorem.
\begin{theorem}[Equivalence of FRS]\label{thm:deter-prob-FRS}
    Let the dynamics $f$ in \eqref{eq:dynamics} be continuously differentiable. Let $\mu_t$ be the push-forward of absolutely continuous probability distributions with support on $\mathcal{U}_i$ and $\mathcal{W}_i$, mapped through the composed dynamics $\hat{f}_{t-1, u_{t-1}, w_{t-1}}\circ \cdots \circ \hat{f}_{0,u_0, w_0}$. 
    Then the sets $F_t$ and $\omega_t^*$ defined in \eqref{eq:deterministic-FRS} and \eqref{eq:probabilistic-FRS} are equal almost everywhere (except for sets with measure zero).
\end{theorem}
The proof of this result is provided in the Appendix.

The probabilistic formulation in \eqref{eq:probabilistic-FRS} offers a powerful bridge to estimating FRS using modern trajectory predictors, which are generative networks that learn distributions $\mu_t$ over future states of agents from some training driving data. 
These models implicitly encode the hard-to-specify uncertainties (e.g., $\mathcal{U}_i$, $\mathcal{W}_i$) and allow us to extract refined, data-driven FRSs that are less conservative than worst-case reachability. Nonetheless, it is worth noting that worst-case approaches remain valuable as fallback strategies when predictors fail due to learning errors or out-of-distribution (OOD) conditions.

%
In the remainder of this paper, we adopt the probabilistic view of the FRS to achieve two key goals:
(i) develop an efficient algorithm, FORCE-OPT, to extract FRSs from learning-based trajectory predictors with probabilistic guarantees of ground-truth coverage; and
(ii) use these calibrated FRSs to build a safety monitor that evaluates motion plans and empirically demonstrate low false-negative and false-positive rates under both in-distribution and OOD operation.

\section{FORCE-OPT}
%
We now present FORCE-OPT, our algorithm for estimating FRS from trajectory predictors and calibrating them using conformal prediction. FORCE-OPT combines learned generative models with convex optimization to efficiently compute calibrated FRSs, and uses Bayesian filtering to hedge against out-of-distribution (OOD) deployment failures.

\subsection{Trajectory Predictors}
\label{subsec:traj-pred}
Without loss of generality, let the current time be $0$. 
Denote the state of all agents in the scene by $\xi \in \mathcal{X}^{n_{\rm agents}}$, and their historical trajectories over a horizon $H$ by $\xi_{-H:0}\in\mathcal{X}^{H n_{\rm agents}}$.
Let $m \in \mathcal{M}$ represent map and other contextual scene information. 
A traffic scene is then denoted by $s:=(\xi_{-H:0}, m)\in \mathcal{X}^H\times\mathcal{M}=:\mathcal{S}$. 

A trajectory predictor is a function $\phi:\mathcal{S} \to \mathcal{P}(\mathcal{X}^T)$ that maps a scene $s\in\mathcal{S}$ to a probability distribution $\hat{\mu}\in\mathcal{P}(\mathcal{X}^T)$ over future trajectories of horizon $T$. 
Most modern trajectory predictors \cite{salzmann20,girgis2021latent} represent $\phi$ using generative neural networks that output a sequence of Gaussian Mixture Models (GMMs) $\{\hat{\mu}_t\}_{t=1}^T$ -- one for each future timestep. 
A GMM with $K$ modes is of the form: $\hat{\mu}_t = \sum_{i=1}^K p_i \hat{\mu}_{t,i}$ where $p_i\geq 0$, $\sum_{i=1}^K p_i = 1$, and each mode $\hat{\mu}_{t,i}$ is a Gaussian distribution $\mathcal{N}(\overline{x}_i, \Sigma_i)$ with mean at $\overline{x}_i$ and a covariance $\Sigma_i$.

\subsection{Extracting FRS from Trajectory Predictors}
\label{subsec:convex-opt}
We treat $\hat{\mu}_t$ as a learned approximation of the push-forward distribution $\mu_t$ in \eqref{eq:probabilistic-FRS}. 
However, given that GMMs have an unbounded support, these measures make it infeasible to find a bounded set $\omega$ satisfying $\hat{\mu}_t(\omega) = 1$. 
To make the problem tractable, we relax the constraint to require only a minimum mass $\tau \in (0, 1)$:
\begin{equation}\label{eq:general-opt}
    \omega^* := \underset{\{\omega\in\Omega: \hat{\mu}_t(\omega) \geq \tau\}}{\arg\inf} \vol(\omega).
\end{equation}

Ideally, we would choose a $\tau$ that is very close to 1. However, at this level of generality, this problem is very challenging to solve. 
Leveraging the fact that the distribution over each timestep is a GMM, we solve a tractable proxy for this problem by restricting the search to unions of sub-level sets of the GMM modes.
For a given mode $\hat{\mu}_{t,i}$, define the Mahalanobis energy function $V_i(x) := (x-\overline{x}_i)^{\rm T} \Sigma_i^{-1} (x-\overline{x}_i)$, and its sublevel set $E_i(c_i) := \{x : V_i(x) \leq c_i\}$ for $c_i \geq 0$. 
We then solve:
\begin{equation}\label{eq:approx-opt}
    \begin{aligned}
        c_1^*, \cdots, c_K^* = & \underset{c_1,\cdots,c_K}{\arg\inf} \sum_{i=1}^K \vol(E_i(c_i)), \\
        & \mathrm{subject~to~} \sum_{i=1}^K p_i \hat{\mu}_{t,i}(E_i) \geq \tau, \\
        & \phantom{\mathrm{subject~to~}} c_i \geq 0 \quad \text{for all } i = 1, \cdots, K.
    \end{aligned}
\end{equation}
The resulting FRS is $E^* = \bigcup_{i=1}^K E_i(c_i^*)$. 
Although $E^*$ is not necessarily the smallest such set, it is a feasible solution to \eqref{eq:general-opt}, and tight in practice. When $\mathcal{X} \subseteq \mathbb{R}^2$, it takes the form of a convex optimization, as detailed in the following theorem.
%
\begin{theorem}[Convex Optimization for FRS Extraction]\label{thm:convex-opt}
    If $\mathcal{X}\subseteq \mathbb{R}^2$, then \eqref{eq:approx-opt} takes the form of a convex optimization:
    \begin{align}\label{eq:convex-opt}
        \begin{aligned}
            c_1^*,\cdots,c_K^* = & \underset{c_1,\cdots,c_K}{\arg\min} \sum_{i=1}^K \pi \sqrt{\lambda_{i,1}\lambda_{i,2}} c_i ,\\
            & \mathrm{subject~to~} \sum_{i=1}^K p_i (1-\exp(-c_i/2)) \geq \tau , \\
            & \phantom{\mathrm{subject~to~}} c_i \geq 0 \quad \text{for all } i = 1, \cdots, K.
        \end{aligned}
    \end{align}
    where $\lambda_{i,1}$ and $\lambda_{i,2}$ are eigen-values of $\Sigma_i$.
\end{theorem}
The proof of this theorem is provided in the Appendix. The optimization in \eqref{eq:convex-opt} has a few key advantages. First, it computes the size of sub-level sets in accordance with the probabilities assigned to individual modes, resulting in ``tighter" sets than if we were to simply choose the $\tau$ probability sub-level sets for each mode of the GMM. 
Second, unlike prior work \cite{tumu2024multi,lew2021sampling}, this approach does not require us to draw samples from the distribution; instead, it exploits the GMM structure to solve the optimization problem and compute the FRS in microseconds, significantly lowering the FRS inference time. 
Finally, the solution to \eqref{eq:convex-opt} is invariant to scaling of the covariance matrix, as shown in the following corollary of Theorem~\ref{thm:convex-opt}. 
This property will be instrumental in conformal calibration of the FRS.
\begin{corollary}[Invariance of \eqref{eq:convex-opt} to Covariance Scaling]\label{cor:scale-invariance}
    The solution of the convex optimization \eqref{eq:convex-opt} established in Theorem~\ref{thm:convex-opt} is the same for any scaling $\alpha\in(0,\infty)$ of the GMM covariances $\{\Sigma_i\}_{i=1}^K$. 
\end{corollary}

\subsection{Conformalizing FRS from Trajectory Predictors}
\label{subsec:CP}
%
%
%
%
The FRS obtained from \eqref{eq:convex-opt} assumes the predictor’s distribution $\hat{\mu}_t$ closely matches the true transition dynamics. In reality, due to modeling choices and limited training data, $\hat{\mu}_t$ may fail to cover the true ground-truth distribution. We address this by applying split conformal prediction to calibrate the reachable set to achieve high-probability coverage of the ground truth future.

We calibrate our FRS by scaling the covariance of the GMM modes. 
To do so, we define a non-conformity score function $\psi(s,x)$ as the smallest scaling factor $\alpha > 0$ such that the ground-truth $x$ lies inside the FRS computed using covariances ${ \alpha \Sigma_i }$:
\begin{equation}
    \psi(s,x):=\inf\{\alpha~|~x\in E^*(\{\alpha\Sigma_i\}_{i=1}^K) \}
\end{equation}
where $s$ is the scene information as defined in Section~\ref{subsec:traj-pred} and $E^*(\{\Sigma_i\}_{i=1}^K)$ be the FRS obtained by solving \eqref{eq:convex-opt} for GMM covariances $\{\Sigma_i\}_{i=1}^K$. Thanks to Corollary~\ref{cor:scale-invariance}, we only need to solve \eqref{eq:convex-opt} once to obtain $c_i^*$, after which we can compute $\psi(s,x)$ analytically: $\psi(s,x) = \min\{V_i(x)/c_i^*~|~i=1,\cdots,K\}$ where $V_i$ is as defined in Section~\ref{subsec:convex-opt} for $\Sigma_i$ that are outputted by the trajectory predictor.

Given $N$ calibration examples $\mathcal{D} := \{(s_j, x_j)\}_{j=1}^N$, if we let $\eta$ be the $(1 - \gamma)$ empirical quantile of the scores ${\psi(s_j, x_j)}$. Then, from \cite[Proposition~2a]{vovk2012conditional}, with probability $1 - \delta$:
%
\begin{align} \label{eqn:conformal_guarantees}
    \Pr_{s,x}\left( x \notin C(s,\eta) \right) < \gamma - \sqrt{-\log\delta / 2 N},
\end{align}
where $C(s,\eta)$ is the conformalized FRS given by:
\begin{align}
    C(s, \eta) &:= \{ x : \psi(s,x) < \eta \} \\ 
    &= \{ x : \inf \{ \alpha | x \in E^*(\{\alpha \Sigma_i\}_{i=1}^K) \} < \eta \}  \\
    &= \bigcup_{0 < \alpha < \eta} E^*(\{\alpha \Sigma_i\}_{i=1}^K) \\
    &= E^*(\{\eta\Sigma_i\}_{i=1}^K)
\end{align}
where the last line follows due to the monotonicity of sub-level sets:
\begin{lemma}\label{lem:monotonic-set-inclusion}
    Let $\alpha_1,\alpha_2 \in (0,\infty)$ and $\alpha_1<\alpha_2$, then $E^*(\{\alpha_1\Sigma_i\}_{i=1}^K) \subseteq E^*(\{\alpha_2\Sigma_i\}_{i=1}^K)$.
\end{lemma}
%
%
Intuitively, \eqref{eqn:conformal_guarantees} states that the probability of ground truth future state lying outside the conformalized FRS $C(s, \eta)$ is very low.

\subsection{Hedging Against OOD Failures via Bayesian Filtering} \label{subsec:bayesian}
The conformalization process in Section~\ref{subsec:CP} allows FORCE-OPT to provide statistical guarantees on the coverage of predicted reachable sets. However, these guarantees hold only under the assumption that test-time inputs are drawn independently and identically (IID) from the same distribution as the calibration data. In practice---especially in autonomous driving---this assumption is often violated, and distribution shift can cause these guarantees to break down.

To address this challenge, we augment FORCE-OPT with a Bayesian filtering mechanism that dynamically adjusts uncertainty in the reachable set when the trajectory predictor appears unreliable. Specifically, we introduce a model confidence parameter $\beta \in [\beta_{\text{low}}, \beta_{\text{high}}]$ that reflects our trust in the predictor. This parameter scales the covariance of the GMMs used in the FRS adaptively, effectively dilating the predicted reachable sets to reflect greater uncertainty.. 

To track this confidence online, we adopt the Bayesian update scheme proposed in \cite{nakamura2023online}. 
At each timestep $t$, we maintain a belief distribution over $\beta$, denoted by $\text{bel}^t(\beta)$. 
The belief is initialized uniformly: $bel^0(\beta_{low}) = bel^0(\beta_{high}) = 0.5$.
The belief is then updated at every timestep using the likelihood of the \textit{observed} agent state $x_t$ under the conformalized GMM, whose covariances are additionally scaled by the inverse of $\beta$:

\vspace{-3mm}
\small
\begin{equation}
\text{bel}^{t+1}(\beta) = 
\frac{\varphi\bigg({x}_{t}; \text{GMM}\bigg(\{\bar{x}^{t}\}_{i=1}^{K}, \frac{1}{\beta} \{\eta\Sigma^{t}_i\}_{i=1}^{K}\bigg) \bigg) \text{bel}^{t}(\beta)}
{\sum_{\tilde{\beta}} \varphi\bigg({x}_{t}; \text{GMM}\bigg(\{\bar{x}^{t}\}_{i=1}^{K}, \frac{1}{\tilde\beta} \{\eta\Sigma^{t}_i\}_{i=1}^{K}\bigg) \bigg) \text{bel}^{t}(\tilde{\beta})},
\end{equation}
\normalsize
where $\tilde\beta \in \{\beta_{\text{low}}, \beta_{\text{high}}\}$. Here $\varphi(x, GMM(\cdot))$ denotes the likelihood of state $x$ under the given GMM. 
Note that the GMM covariances are already scaled by the conformal calibration factor $\eta$; the inverse $\beta$ scaling further adjusts the uncertainty based on current trust in the predictor.
Finally, we compute the effective model confidence as the expected value under the belief: $\hat\beta = \mathbb{E}[\beta]$, and use this $\hat{\beta}$ to further scale the covariances of the distribution to adjust the final FRS returned by FORCE-OPT.

While conformal prediction provides statistical guarantees on coverage, these guarantees hold only in the average case. It does not account for rare but high-impact failures that may lie in the tails of the distribution. To address such worst-case scenarios, we incorporate techniques from Hamilton-Jacobi (HJ) reachability analysis. Specifically, when the online model confidence $\beta$ drops below a critical threshold, it indicates that the predictor may be operating outside the distribution represented by the calibration dataset. These are precisely the conditions under which FORCE-OPT becomes vulnerable to long-tailed failure modes.

To mitigate this risk, we employ two fallback mechanisms based on reachability theory: \textbf{(i)} a Parameterized FRS that adapts to observed uncertainty levels, and \textbf{(ii)} a Worst-Case FRS that assumes bounded adversarial disturbances. These fallback strategies are activated when $\beta$ falls below the specified threshold. We explore the behavior and trade-offs of these approaches in detail in the results section, particularly in the context of different predictor types.

The belief update mechanism and switching strategy allow FORCE-OPT to maintain tight, calibrated bounds in-distribution, while conservatively hedging against uncertainty in out-of-distribution or low-confidence scenarios.

\section{Experimental Results}

\label{sec:result}

We now present an empirical evaluation of FORCEOPT to assess its effectiveness in safety-critical motion planning. Our experiments are designed to address the following key research questions: \qa{(Q1)} How effective is FORCE-OPT at balancing completeness and soundness of safety evaluation for autonomous driving? \qa{(Q2)} Do the belief-based extensions of FORCE-OPT result in more robust monitoring in out-of-distribution operation? \qa{(Q3)} How effectively can FORCE-OPT and its extensions leverage multimodality of trajectory prediction? \qa{(Q4)} Does FORCE-OPT offer a computational advantage that might enable online deployment?

    
    


\subsection{Datasets}
To answer the above questions, we evaluate FORCE-OPT's performance on nuScenes \cite{nuscenes}, a large-scale autonomous driving dataset. The nuScenes dataset includes approximately 15 hours of expert-labeled driving data in Boston and Singapore. 
%
We train the Autobots predictor \cite{girgis2021latent} on the training split of Singapore and then test it on the test splits of both cities. This setup allows us to test the performance of our approach on in distribution (ID) where the model is trained and tested on the same city as well as out-of-distribution (OOD) where the model is tested on a city different from the one it is trained on. 


\subsection{Synthetic Unsafe Data Generation}
Although nuScenes provides diverse testing conditions, all the driving data available in the dataset is inherently safe, which makes it challenging to assess the monitor's ability to detect potential safety violations.
To remedy this, we synthetically generate unsafe scenarios by modifying scenes within the nuScenes dataset. 
Specifically, we identify potential intersections between the trajectories of the ego vehicle and a surrounding agent, defining the point of closest approach $p_c$ as the collision point. 
Let the ego and the contender arrive at $p_c$ at times $t_e$ and $t_o$, respectively.
We use a bicycle dynamics model for the ego vehicle and optimize its trajectory using Sequential Least Squares Programming (SLSQP) to force it to reach $p_c$ at $t_o$, while obeying initial conditions and physical constraints.
The resulting trajectory of the ego vehicle, along with the original trajectory of the contender, constitutes a synthesized unsafe scenario.
This yields plausible but unsafe plans, which serve as ground truth for evaluating false negative rates.

\subsection{Metrics}
The primary objective of this paper is to assess the safety of motion plans. To that end, we focus on two broad categories of metrics: \textbf{completeness} and \textbf{soundness}. These metrics evaluate on a per-frame basis how effectively and reliably a safety assessment algorithm captures true safety violations while minimizing over-conservatism.
\begin{itemize}
    \item \textbf{Completeness} is quantified using two key metrics:                     \begin{enumerate} 
                \item \textit{Coverage (Cov):} 
                The fraction of ground-truth future trajectories that fall within the predicted reachable set.
                High coverage indicates that the predicted FRS captures the actual future behavior well.
                \item \textit{False Negative Rate (FNR):} 
                The proportion of true collision cases (from synthesized unsafe data) that are not flagged by the monitor. 
            \end{enumerate}
    \item \textbf{Soundness} is evaluated using the \textit{False Positive Rate (FPR)}: The fraction of safe scenarios that are incorrectly flagged as unsafe. A low FPR indicates soundness, avoiding unnecessary interventions.
    \item \textbf{Balance between Completeness and Soundness} is evaluated using the \emph{Balanced Error Rate (BER)}: Arithmetic mean of FPR and FNR effectively capturing the tradeoff between them.
\end{itemize}

\begin{table*}[ht]
\centering
\setlength{\tabcolsep}{2pt}
\small
\resizebox{1.0\textwidth}{!}{
\begin{tabular}{c|ccc|ccccc|c}
\toprule
\textbf{Approach} & \multicolumn{3}{c|}{\textbf{Uncalibrated Trajectory Predictor}} & \multicolumn{5}{c|}{\textbf{Calibrated Trajectory Predictor with Conformal Prediction}} & \textbf{Data-free} \\
\midrule
\textbf{Metric} & 99\% & Parametric & Nakamura & Lindemann & FORCE-OPT & FORCE-OPT+ & FORCE-OPT+ & FORCE-OPT+ & Worst Case \\
~& CI & WC-FRS & et al. \cite{nakamura2023online} & et al. \cite{lindemann2023safe} & (ours) & belief (ours) & pWC-FRS (ours) & WC-FRS (ours) & FRS\\
\midrule
\textbf{Cov} $(\uparrow)$ & 54.76\% & 97.40\% & 86.84\% & 89.50\% & 89.95\% & 93.94\% & 94.90\% & 96.01\% & \textbf{99.53\%} \\
\textbf{FPR} $(\downarrow)$ & \textbf{1.19\%} & 19.68\% & 7.33\% & 31.12\% & 8.62\% & 15.57\% & 13.35\% & 24.43\% & 43.73\% \\
\textbf{FNR} $(\downarrow)$ & 33.33\% & \textbf{0\%} & 36.36\% & \textbf{0\%} & 3.03\% & \textbf{0\%} & \textbf{0\%} & \textbf{0\%} & \textbf{0\%} \\
\textbf{BER} $(\downarrow)$ & 17.26\% & 9.84\% & 20.84\% & 15.56\% & \textbf{5.83\%} & 7.78\% & 6.78\% & 12.21\% & 21.86\% \\
\bottomrule
\end{tabular}
}
\caption{\small{Performance comparison across methods (best-performing values in bold) in Singapore (ID).}}
\label{tab:method_comparison}
\end{table*}
\vspace{-1em}

\subsection{Baselines and Variants}
\label{subsec:baselines}
We categorize all methods in three broad categories:
\begin{itemize}
    \item \textbf{Uncalibrated Trajectory Predictor:} The methods that fall under this class directly use the trajectory predictor without calibrating them. \textbf{99\% CI}: The original predictor where the sets occupy 99\% of GMM probability mass. $E = \bigcup_{i=1}^K E_i(C)$ where C = value at 99th Percentile of a $\chi^2$ distribution. \textbf{Parametric Worst Case-FRS (pWC-FRS)}: We obtain the control bounds as the $3\sigma$ (or 99\% confidence interval) support of the Gaussian control distribution for each mode predicted by a trajectory predictor and then estimate a worst-case FRS for each mode and take the union of the sets. (the parameter is the velocity and the control bound of he agent) \textbf{Nakamura et al. \cite{nakamura2023online}}: Adapts FRS via belief tracking of a trajectory predictor's performance. In this case the control bounds enclose the 3\% probaility mass around the mean of the normal distribution predicted by the trajectory predictor  \cite{salzmann20}.
    
    \item \textbf{Calibrated Trajectory Predictor:} The methods within this class leverage trajectory predictors that are calibrated using CP. We use a dataset with a cardinality of 35220 for calibration and set the desired coverage probability in CP to 0.95.  \textbf{Lindemann et al.} \cite{lindemann2023safe}: Provides coverage guarantees via conformalization between the highest likelihood trajectory and the ground truth.  \textbf{FORCE-OPT (Ours)}: Solves the convex optimization from Sec.~\ref{subsec:convex-opt} using GMMs from learned predictors, along with the calibration schemes in Sec.\ref{subsec:CP}. \textbf{FORCE-OPT + belief}: Additionally adjusts the GMM covariances of FORCE-OPT using Bayesian filtering approach in Sec. \ref{subsec:bayesian}. For this and all the subsequent methods that use belief-based adaptation, we set $\beta_{\rm low} = 0.3$ and $\beta_{\rm high} = 1$. \textbf{FORCE-OPT + pWC-FRS}: A hybrid approach that switches from FORCE-OPT to Parametric WC-FRS when $\beta<0.75$ indicating a drop in the predictor's performance. \textbf{FORCE-OPT + WC-FRS}: This approach switches from FORCE-OPT to worst-case FRS when $\beta<0.75$.
    
    \item \textbf{Data-Free:} This category contains only one baseline metric \textbf{Worst Case FRS (WC-FRS)} which computes worst-case FRS assuming 4D Dubins vehicle dynamics with bounded control inputs.
\end{itemize}

\subsection{Results and Discussion}

The results for in-distribution evaluation of all approaches discussed in Section~\ref{subsec:baselines} are summarized in Table \ref{tab:method_comparison}, while those for OOD are summarized in Table~\ref{tab:method_comparison_OOD}.

\subsubsection{Balance between Completeness and Soundness \qa{(Q1)}}
\label{subsubsec:balance}

From Table~\ref{tab:method_comparison} we observe that FORCE-OPT achieves the lowest BER, indicating the best balance between FPR and FNR. The metrics that use CP-based calibration on the trajectory predictor have a significantly lower FNR at the expense of a slightly higher FPR than the ones that do not calibrate the predictor, generally indicating that the trajectory predictor without calibration tends to be over-optimistic in safety assessment missing out safety critical events, highlighting the importance of calibrating the trajectory predictor---one exception to this observation is the parametric FRS that counters the uncalibrated predictor's over-optimism by plugging the predicted control bounds to solve for the worst-case FRS. We also note that FORCE-OPT and its belief-based extensions, in comparison to \cite{lindemann2023safe}, have a lower FPR while maintaining a similar FNR and higher coverage rates. These additional gains are the outcome of the Theorem~\ref{thm:deter-prob-FRS}-inspired convex optimization \eqref{eq:convex-opt} that more precisely models the forward reachable space while leveraging multi-modality (unlike \cite{lindemann2023safe} which uses a single mode); this is discussed in greater detail in Section~\ref{subsubsec:multimodality}. FORCE-OPT has lower FPR and slightly worse FNR than its belief-based variants, which is in alignment with our expectations as the belief-based approaches introduce greater conservatism in the event when the predictor's performance deteriorates.

\begin{table*}[ht]
\centering
\setlength{\tabcolsep}{2pt}
\small
\resizebox{1.0\textwidth}{!}{
\begin{tabular}{c|ccc|ccccc|c}
\toprule
\textbf{Approach} & \multicolumn{3}{c|}{\textbf{Uncalibrated Trajectory Predictor}} & \multicolumn{5}{c|}{\textbf{Calibrated Trajectory Predictor with Conformal Prediction}} & \textbf{Data-free} \\
\midrule
\textbf{Metric} & 99\% & Parametric & Nakamura & Lindemann & FORCE-OPT & FORCE-OPT+ & FORCE-OPT+ & FORCE-OPT+ & Worst Case \\
~& CI & WC-FRS & et al. \cite{nakamura2023online} & et al. \cite{lindemann2023safe} & (ours) & belief (ours) & pWC-FRS (ours) & WC-FRS (ours) & FRS\\
\midrule
\textbf{Cov} $(\uparrow)$ & 62.85\% & 97.45\% & 88.99\% & 88.02\% & 90.06\% & 93.57\% & 95.07\% & 96.06\% & \textbf{99.46\%} \\
\textbf{FPR} $(\downarrow)$ & \textbf{1.18\%} &17.08\% & 7.27\% & 36.38\% & 5.22\% & 9.16\% & 9.85\% & 21.15\% & 45.42\% \\
\textbf{FNR} $(\downarrow)$ & 51.85\% & 3.70\% & 55.56\% & 9.26\% & 12.96\% & 7.41\% & 3.70\% &  \textbf{0\%} & \textbf{0\%} \\
\textbf{BER} $(\downarrow)$ & 26.52\% & 10.39\% & 31.41\% & 22.79\% &  9.09\%& 8.28\% & \textbf{6.77}\% & 10.57\% & 22.71\% \\
\bottomrule
\end{tabular}
}
\caption{\small{Performance comparison across methods (best-performing values in bold) in Boston (OOD).}}
\label{tab:method_comparison_OOD}
\end{table*}
\vspace{-1em}

\subsubsection{Out-of-distribution Robustness \qa{(Q2)}}

The belief-based adaptation mechanism enables smooth adjustment to uncertain contexts without excessive conservatism, as evidenced by the fact that the performance of the belief-based variants of FORCE-OPT remains similar between Tables~\ref{tab:method_comparison} and \ref{tab:method_comparison_OOD} and the best performing metric according to BER in Table~\ref{tab:method_comparison_OOD} is FORCE-OPT + pWC-FRS. Comparing the results in Table~\ref{tab:method_comparison} (ID) and Table~\ref{tab:method_comparison_OOD} (OOD), we also observe that the metrics that use an uncalibrated trajectory predictor suffer a drop in performance, especially with the FNR which jumps from 33.33\% and 36.36\% (ID) to 51.85\% and 55.56\% (OOD) for 99\% CI and \cite{nakamura2023online}, respectively---as before, parametric FRS is an exception to this for the same reason mentioned in Section~\ref{subsubsec:balance}. On the other hand, the approaches that use the calibrated trajectory predictor exhibit stronger OOD robustness; albeit, there is a modest increase in the FNR for all these approaches. Unsurprisingly, the performance of Worst-Case FRS is unaffected by the distribution shift.


\subsubsection{Impact of Multi-modality \qa{(Q3)}}
\label{subsubsec:multimodality}

\begin{figure*}[t]
    \centering
    \begin{subfigure}[b]{0.33\textwidth}
        \includegraphics[width=\textwidth]{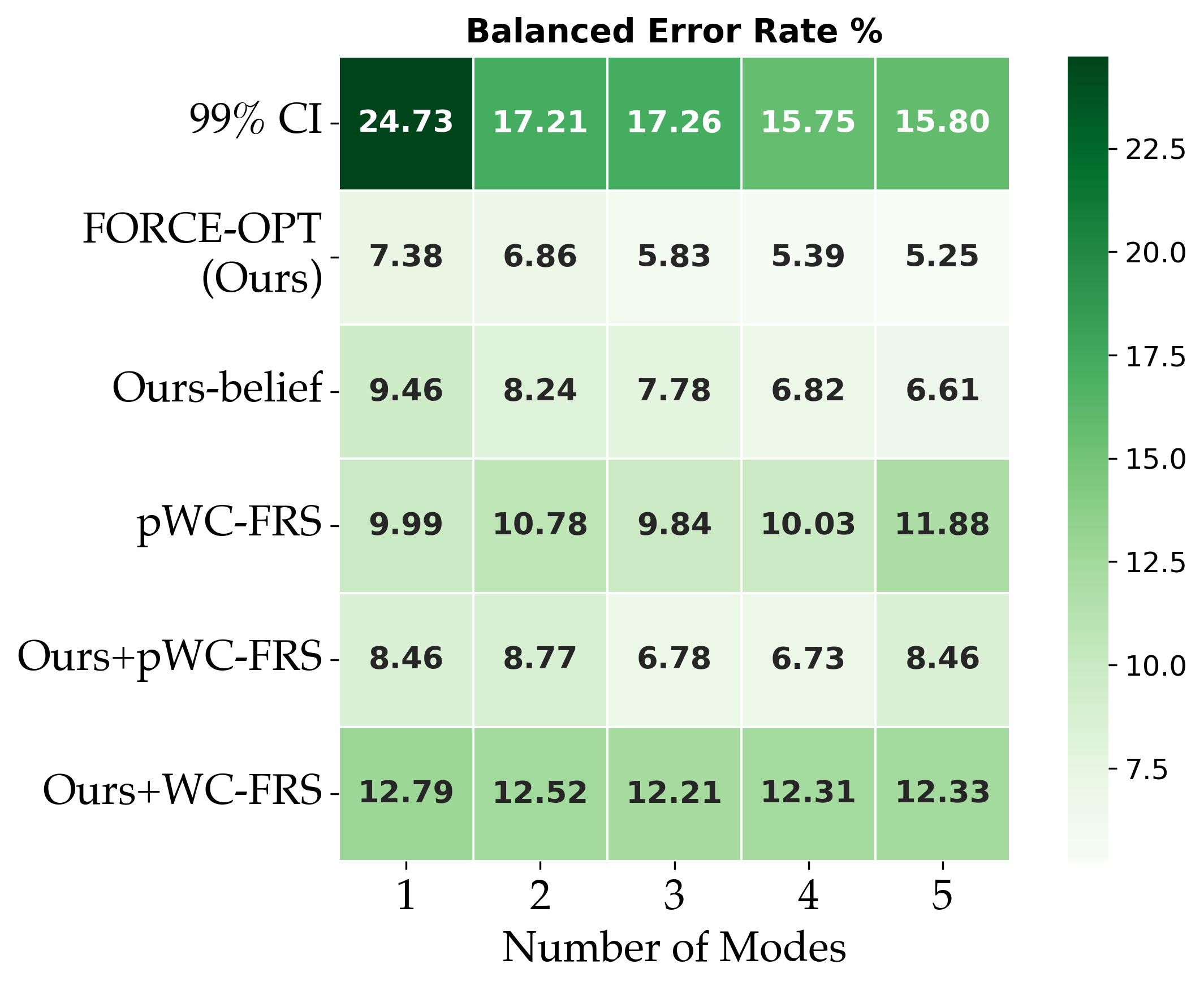}
        \caption{\small{BER vs. Number of Modes}}
        \label{fig:num_modes_ber}
    \end{subfigure}
    \hspace{-5mm}
    \hfill
    \begin{subfigure}[b]{0.33\textwidth}
        \includegraphics[width=\textwidth]{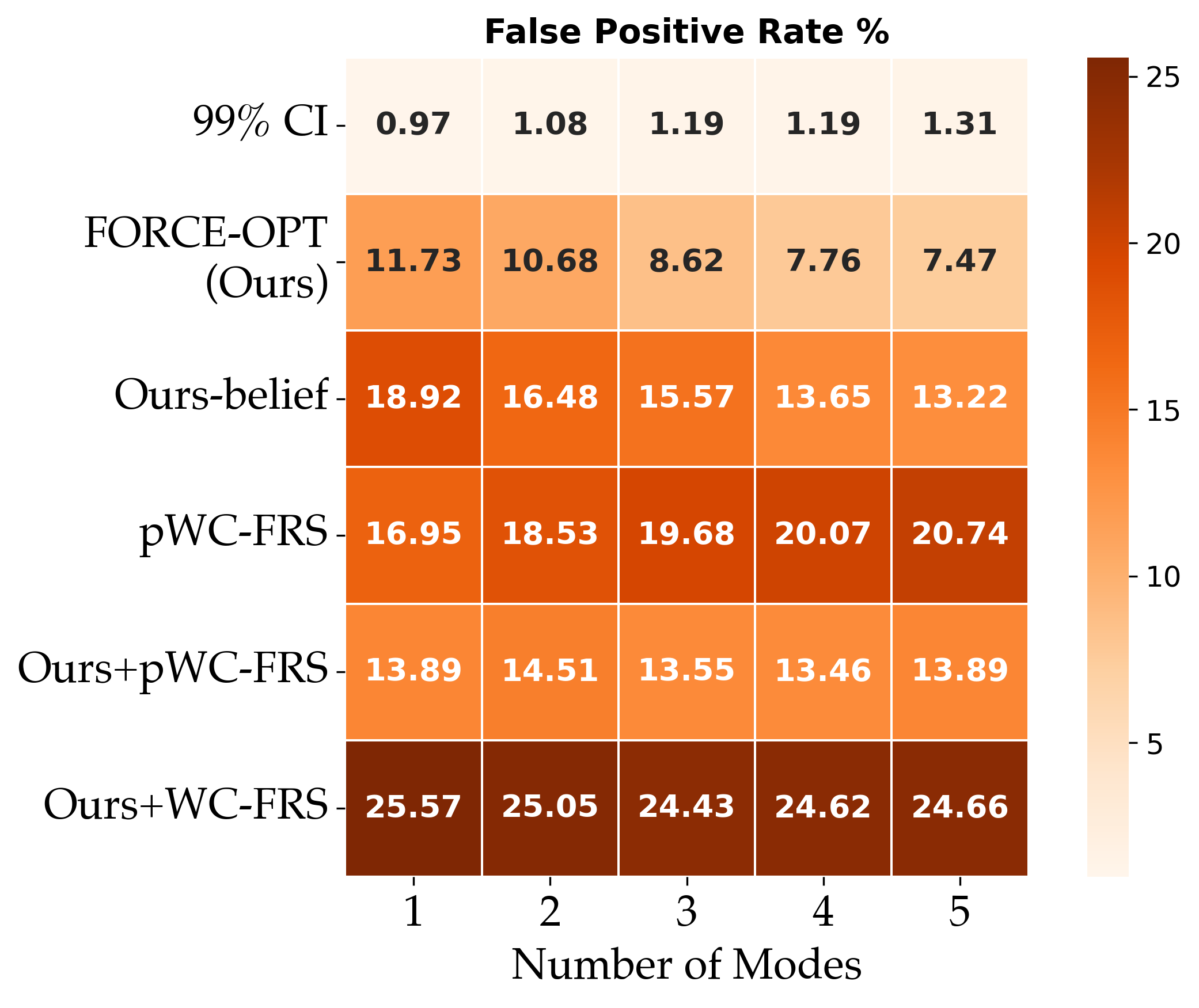}
        \caption{\small{FPR vs. Number of Modes}}
        \label{fig:num_modes_fpr}
    \end{subfigure}
    \hspace{-5mm}
    \hfill
    \begin{subfigure}[b]{0.33\textwidth}
        \includegraphics[width=\textwidth]{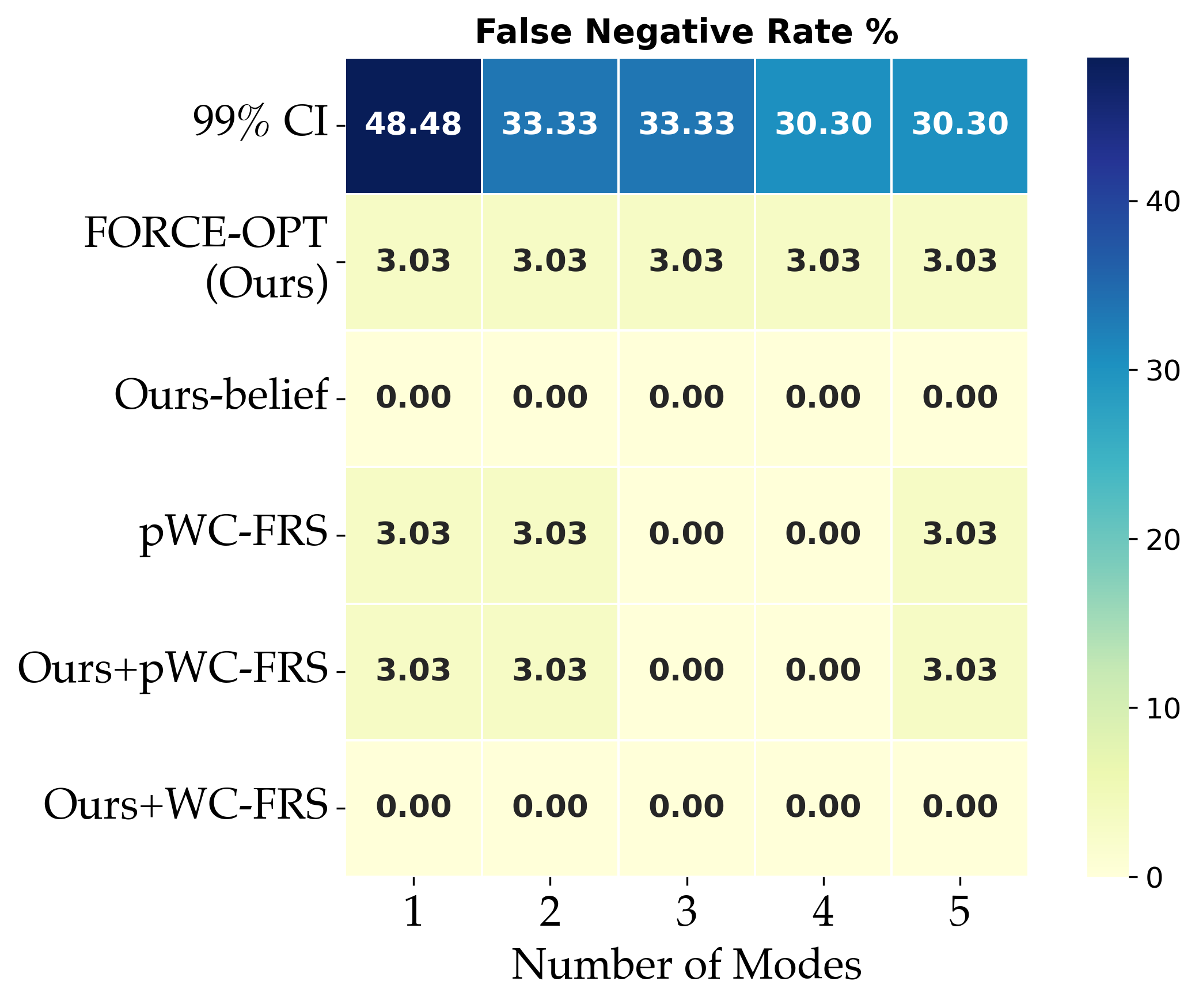}
        \caption{\small{FNR vs. Number of Modes}}
        \label{fig:num_modes_fnr}
    \end{subfigure}
    \caption{\small{Ablation with different number of GMM modes from the trajectory predictor. The performance of FORCE-OPT and its belief-based variants improves with more GMM modes.}}
    \label{fig:num_modes}
\end{figure*}

One of the key strengths of FORCE-OPT is its ability to systematically integrate probabilities from multiple prediction modes. To study the impact of multi-modality, we ablate the performance of the metrics in Section~\ref{subsec:baselines} that can handle multimodality, i.e., 99\% CI, parametric FRS, FORCE-OPT and its belief-based variants, on the Singapore dataset against the number of GMM modes. As the number of modes increase from one to five, we observe that the BER drops for all methods other than pWC-FRS Fig.~\ref{fig:num_modes_ber}. The drop in BER in different methods is fueled by different reasons: for 99\% CI, the BER improves because the FNR improves with more modes Fig.~\ref{fig:num_modes_fnr}, while for FORCE-OPT and its variants the BER improvement arises from an improvement in the FPR Fig.~\ref{fig:num_modes_fpr}. At first look, it is indeed surprising that more modes result in a lower FPR for FORCE-OPT; however, this counter-intuitive outcome is the result of the fact that greater multi-modality requires smaller set inflations via CP, as evidenced by the fact that the $\alpha$'s decrease as the number of modes increase, as shown in Table~\ref{tab:alphas}. If a mode other than the most-likely one is nearer to the ground truth in the calibration set, then the amount of set inflation needed to cover that ground-truth position would have to be less than the inflation needed for the most-likely mode that is further away. Overall, our ablations suggest that greater multi-modality promotes better safety assessment by allowing us to reason about multiple plausible future outcomes which could be closer to the ground truth behavior than whatever the model deems to be the most likely.

\begin{table}[h!]
\centering
\setlength{\tabcolsep}{2pt}
\small
\begin{tabular}{c|cccccc}
\toprule
\textbf{\# Modes} & $\mathbf{t=1}$ & $\mathbf{t=2}$ & $\mathbf{t=3}$ & $\mathbf{t=4}$ & $\mathbf{t=5}$ & $\mathbf{t=6}$ \\
\midrule
1 & 3.18  & 11.73 & 21.63 & 32.84 & 43.99 & 53.44 \\
2 & 2.23  & 7.72  & 14.51 & 23.36 & 32.3  & 41.43 \\
3 & 2.11  & 7.07  & 13.21 & 21.06 & 29.23 & 37.48 \\
4 & 2.04  & 6.45  & 11.67 & 18.69 & 25.55 & 33.65 \\
5 & 1.92  & 6.10  & 11.20 & 17.43 & 24.05 & 31.72 \\
\hline
\end{tabular}
\caption{\small{Calibrated $\alpha$ for each timestep along the predicted trajectory as the number of modes vary. The table shows that as the number of modes increase the amount of set inflation $\alpha$ needed for calibration reduces.}}
\label{tab:alphas}
\end{table}
\vspace{-1em}

%
%

\subsubsection{Compuational Efficiency \qa{(Q4)}}

In Table \ref{tab:timing}, we show the computation time for all the methods presented in Section~\ref{subsec:baselines} along with their performance on BER in Tables~\ref{tab:method_comparison} and \ref{tab:method_comparison_OOD}. FORCE-OPT demonstrates fast runtimes while achieving strong BER results in both ID and OOD settings, outperforming faster baselines such as 99\% CI and \cite{lindemann2023safe}. Notably, adding belief tracking adds negligible overhead—FORCE-OPT + belief is only 0.001 seconds slower than FORCE-OPT alone. With the exception of FORCE-OPT + pWC-FRS, all other variants of FORCE-OPT are faster than 0.1 seconds, suggesting that these algorithms are well-suited for deployment in real-time, safety-critical applications. 


\begin{table}[ht]
\centering
\setlength{\tabcolsep}{2pt}
\resizebox{0.95\columnwidth}{!}{
\begin{tabular}{c c c c}
\toprule
\textbf{Method} & \textbf{Time (s)} & \textbf{ID BER} & \textbf{OOD BER} \\
\midrule
99\% CI & \textbf{0.001} & 17.26\% & 26.52\% \\
pWC-FRS & 0.187 & 9.84\% & 10.39\% \\
Nakamura et al. \cite{nakamura2023online} & 0.155 & 20.84\% & 31.41\% \\
Lindemann et al. \cite{lindemann2023safe} & 0.005 & 15.56\% & 22.79\% \\
FORCE-OPT (ours) & 0.022 & \textbf{5.83\%} & 9.09\% \\
FORCE-OPT + belief (ours) & 0.023 & 7.78\% & 8.28\% \\
FORCE-OPT + pWC-FRS (ours) & 0.210 & 6.78\% & \textbf{6.77\%} \\
FORCE-OPT + WC-FRS (ours) & 0.079 & 12.21\% & 10.57\% \\
WC-FRS & 0.056 & 21.86\% & 22.71\% \\
\bottomrule
\end{tabular}
}
\caption{Computation Time for Different FRS Methods} 
\normalsize
\label{tab:timing}
\end{table}
\vspace{-1em}

\section{Conclusion and Future Work}
This paper introduced FORCE-OPT, a principled framework for evaluating the safety of motion plans using trajectory predictors as estimators of forward reachable sets. By combining convex optimization, conformal prediction, and Bayesian filtering, our method generates calibrated uncertainty sets that balance completeness (low false negatives) with soundness (low false positives). Empirical results on nuScenes demonstrate that FORCE-OPT significantly outperforms both conservative model-based and raw learning-based baselines, while gracefully handling out-of-distribution scenarios. We believe FORCE-OPT offers a promising building block for runtime safety monitoring in learned autonomy stacks.

4This work opens up several directions for exploration: (i) While the trajectory predictor conditions on scene context, FORCE-OPT itself operates independently for each agent when computing FRS. Joint multi-agent reachability, especially in dense traffic scenarios with interdependent behaviors, remains an open direction. (ii) Trajectory predictors are trained to distributionally mimic the observed data, not to facilitate the extraction of FRS. Training a neural FRS generator that directly outputs sets is another exciting open direction.




\bibliographystyle{IEEEtran}
\bibliography{references}

\begin{thebibliography}{10}
\providecommand{\url}[1]{#1}
\csname url@samestyle\endcsname
\providecommand{\newblock}{\relax}
\providecommand{\bibinfo}[2]{#2}
\providecommand{\BIBentrySTDinterwordspacing}{\spaceskip=0pt\relax}
\providecommand{\BIBentryALTinterwordstretchfactor}{4}
\providecommand{\BIBentryALTinterwordspacing}{\spaceskip=\fontdimen2\font plus
\BIBentryALTinterwordstretchfactor\fontdimen3\font minus \fontdimen4\font\relax}
\providecommand{\BIBforeignlanguage}[2]{{%
\expandafter\ifx\csname l@#1\endcsname\relax
\typeout{** WARNING: IEEEtran.bst: No hyphenation pattern has been}%
\typeout{** loaded for the language `#1'. Using the pattern for}%
\typeout{** the default language instead.}%
\else
\language=\csname l@#1\endcsname
\fi
#2}}
\providecommand{\BIBdecl}{\relax}
\BIBdecl

\bibitem{teng2023motion}
S.~Teng \emph{et~al.}, ``Motion planning for autonomous driving: The state of the art and future perspectives,'' \emph{TIV}, vol.~8, no.~6, pp. 3692--3711, 2023.

\bibitem{schwarting2018planning}
W.~Schwarting \emph{et~al.}, ``Planning and decision-making for autonomous vehicles,'' \emph{Annual Review of CRAS}, vol.~1, no.~1, pp. 187--210, 2018.

\bibitem{salzmann2020trajectron++}
T.~Salzmann \emph{et~al.}, ``Trajectron++: Dynamically-feasible trajectory forecasting with heterogeneous data,'' in \emph{ECCV}.\hskip 1em plus 0.5em minus 0.4em\relax Springer, 2020, pp. 683--700.

\bibitem{girgis2021latent}
R.~Girgis \emph{et~al.}, ``Latent variable sequential set transformers for joint multi-agent motion prediction,'' \emph{arXiv preprint arXiv:2104.00563}, 2021.

\bibitem{angelopoulos2021gentle}
A.~N. Angelopoulos and S.~Bates, ``A gentle introduction to conformal prediction and distribution-free uncertainty quantification,'' \emph{arXiv preprint arXiv:2107.07511}, 2021.

\bibitem{shafer2008tutorial}
G.~Shafer and V.~Vovk, ``A tutorial on conformal prediction.'' \emph{JMLR}, vol.~9, no.~3, 2008.

\bibitem{nuscenes}
H.~Caesar \emph{et~al.}, ``nuscenes: A multimodal dataset for autonomous driving,'' 2020, pp. 11\,621--11\,631.

\bibitem{luckcuck2019formal}
M.~Luckcuck \emph{et~al.}, ``Formal specification and verification of autonomous robotic systems: A survey,'' \emph{CSUR}, vol.~52, no.~5, pp. 1--41, 2019.

\bibitem{mitra2024formal}
S.~Mitra \emph{et~al.}, ``Formal verification techniques for vision-based autonomous systems--a survey,'' in \emph{Principles of Verification: Cycling the Probabilistic Landscape: Essays Dedicated to Joost-Pieter Katoen on the Occasion of His 60th Birthday, Part III}.\hskip 1em plus 0.5em minus 0.4em\relax Springer, 2024, pp. 89--108.

\bibitem{althoff2011zonotope}
M.~Althoff and J.~M. Dolan, ``Reachability analysis of nonlinear systems with uncertain parameters using conservative linearization,'' \emph{CDC}, 2011.

\bibitem{lygeros1999controllers}
J.~Lygeros \emph{et~al.}, ``Controllers for reachability specifications for hybrid systems,'' \emph{Automatica}, vol.~35, no.~3, pp. 349--370, 1999.

\bibitem{mitchell2005toolbox}
I.~M. Mitchell \emph{et~al.}, ``A toolbox of hamilton–jacobi solvers for analysis of nondeterministic continuous and hybrid systems,'' \emph{Hybrid Systems: Computation and Control}, 2005.

\bibitem{majumdar2017funnel}
A.~Majumdar and R.~Tedrake, ``Funnel libraries for real-time robust feedback motion planning,'' \emph{IJRR}, vol.~36, no.~8, pp. 947--982, 2017.

\bibitem{fridovich2020confidence}
D.~Fridovich-Keil \emph{et~al.}, ``Confidence-aware motion prediction for real-time collision avoidance1,'' \emph{IJRR}, vol.~39, no. 2-3, pp. 250--265, 2020.

\bibitem{salzmann20}
T.~Salzmann \emph{et~al.}, ``Trajectron++: Dynamically-feasible trajectory forecasting with heterogeneous data,'' in \emph{ECCV}, 2020, pp. 683--700.

\bibitem{hwang2024emma}
J.-J. Hwang \emph{et~al.}, ``Emma: End-to-end multimodal model for autonomous driving,'' \emph{arXiv preprint arXiv:2410.23262}, 2024.

\bibitem{karkus2023diffstack}
P.~Karkus \emph{et~al.}, ``Diffstack: A differentiable and modular control stack for autonomous vehicles,'' in \emph{CoRL}.\hskip 1em plus 0.5em minus 0.4em\relax PMLR, 2023, pp. 2170--2180.

\bibitem{chen2023interactive}
Y.~Chen \emph{et~al.}, ``Interactive joint planning for autonomous vehicles,'' \emph{RA-L}, 2023.

\bibitem{casas2021mp3}
S.~Casas \emph{et~al.}, ``Mp3: A unified model to map, perceive, predict and plan,'' in \emph{CVPR}, 2021, pp. 14\,403--14\,412.

\bibitem{cui2021lookout}
A.~Cui \emph{et~al.}, ``Lookout: Diverse multi-future prediction and planning for self-driving,'' in \emph{ICCV}, 2021, pp. 16\,107--16\,116.

\bibitem{antonante2023task}
P.~Antonante \emph{et~al.}, ``Task-aware risk estimation of perception failures for autonomous vehicles,'' \emph{arXiv preprint arXiv:2305.01870}, 2023.

\bibitem{chakraborty2024system}
K.~Chakraborty \emph{et~al.}, ``System-level safety monitoring and recovery for perception failures in autonomous vehicles,'' \emph{arXiv preprint arXiv:2409.17630}, 2024.

\bibitem{dinparastdjadid2023measuring}
A.~Dinparastdjadid \emph{et~al.}, ``Measuring surprise in the wild,'' \emph{arXiv preprint arXiv:2305.07733}, 2023.

\bibitem{ding2025surprise}
W.~Ding \emph{et~al.}, ``Surprise potential as a measure of interactivity in driving scenarios,'' \emph{arXiv preprint arXiv:2502.05677}, 2025.

\bibitem{nakamura2023online}
K.~Nakamura and S.~Bansal, ``Online update of safety assurances using confidence-based predictions,'' in \emph{ICRA}.\hskip 1em plus 0.5em minus 0.4em\relax IEEE, 2023, pp. 12\,765--12\,771.

\bibitem{lindemann2023safe}
L.~Lindemann \emph{et~al.}, ``Safe planning in dynamic environments using conformal prediction,'' \emph{RA-L}, 2023.

\bibitem{li2020prediction}
A.~Li \emph{et~al.}, ``Prediction-based reachability for collision avoidance in autonomous driving,'' \emph{arXiv preprint arXiv:2011.12406}, 2020.

\bibitem{althoff2021set}
M.~Althoff \emph{et~al.}, ``Set propagation techniques for reachability analysis,'' \emph{Annual Review of CRAS}, vol.~4, no.~1, pp. 369--395, 2021.

\bibitem{paparusso2024zapp}
L.~Paparusso \emph{et~al.}, ``Zapp! zonotope agreement of prediction and planning for continuous-time collision avoidance with discrete-time dynamics,'' in \emph{ICRA}.\hskip 1em plus 0.5em minus 0.4em\relax IEEE, 2024, pp. 9285--9292.

\bibitem{dixit2023adaptive}
A.~Dixit \emph{et~al.}, ``Adaptive conformal prediction for motion planning among dynamic agents,'' in \emph{L4DC}.\hskip 1em plus 0.5em minus 0.4em\relax PMLR, 2023, pp. 300--314.

\bibitem{chen2021reactive}
Y.~Chen \emph{et~al.}, ``Reactive motion planning with probabilisticsafety guarantees,'' in \emph{CoRL}.\hskip 1em plus 0.5em minus 0.4em\relax PMLR, 2021, pp. 1958--1970.

\bibitem{driggs2017robust}
K.~Driggs-Campbell \emph{et~al.}, ``Robust, informative human-in-the-loop predictions via empirical reachable sets,'' \emph{arXiv preprint arXiv:1705.00748}, 2017.

\bibitem{devonport2020estimating}
A.~Devonport and M.~Arcak, ``Estimating reachable sets with scenario optimization,'' in \emph{L4DC}.\hskip 1em plus 0.5em minus 0.4em\relax PMLR, 2020, pp. 75--84.

\bibitem{tumu2024multi}
R.~Tumu \emph{et~al.}, ``Multi-modal conformal prediction regions by optimizing convex shape templates,'' in \emph{L4DC}.\hskip 1em plus 0.5em minus 0.4em\relax PMLR, 2024, pp. 1343--1356.

\bibitem{xiang2024convex}
J.~Xiang and J.~Chen, ``Convex approximation of probabilistic reachable sets from small samples using self-supervised neural networks,'' \emph{arXiv preprint arXiv:2411.14356}, 2024.

\bibitem{dietrich2025data}
E.~Dietrich \emph{et~al.}, ``Data-driven reachability with scenario optimization and the holdout method,'' \emph{arXiv preprint arXiv:2504.06541}, 2025.

\bibitem{driggs2017integrating}
K.~Driggs-Campbell \emph{et~al.}, ``Integrating intuitive driver models in autonomous planning for interactive maneuvers,'' \emph{ITS}, vol.~18, no.~12, pp. 3461--3472, 2017.

\bibitem{lew2021sampling}
T.~Lew and M.~Pavone, ``Sampling-based reachability analysis: A random set theory approach with adversarial sampling,'' in \emph{CoRL}.\hskip 1em plus 0.5em minus 0.4em\relax PMLR, 2021, pp. 2055--2070.

\bibitem{vovk2012conditional}
V.~Vovk, ``Conditional validity of inductive conformal predictors,'' in \emph{ACML}.\hskip 1em plus 0.5em minus 0.4em\relax PMLR, 2012, pp. 475--490.

\bibitem{wilson2010volume}
A.~J. Wilson, ``Volume of n-dimensional ellipsoid,'' \emph{Sciencia Acta Xaveriana}, vol.~1, no.~1, pp. 101--6, 2010.

\bibitem{zorin1989distribution}
V.~Zorin \emph{et~al.}, ``The distribution of quadratic forms of gaussian vectors,'' \emph{Theory of Probability \& Its Applications}, vol.~33, no.~3, pp. 557--560, 1989.

\end{thebibliography}

\newpage
\section*{APPENDIX}

\section*{Proofs}

\begin{lemma}\label{lem:subset}
    Let $A$ and $B$ be measurable sets. If every measurable subset of $A$ with non-zero measure is a subset of $B$, and if every measurable subset of $B$ with non-zero measure is a subset of $A$, then $A=B$ almost everywhere.
\end{lemma}
\begin{proof}
    We will prove this by contradiction. Let's assume that the hypothesis of the lemma holds, but $\vol(A\setminus B) > 0$. Since $A\setminus B=A \cap B^{\rm c}$, it is a measurable set and $A\setminus B \subseteq A$, i.e., it is a measurable subset of $A$. By the hypothesis of the lemma, we have that $A\setminus B \subseteq B$ (as every measurable subset of $A$ with non-zero measure is a subset of $B$). However, by definition, $A\setminus B$ is disjoint from $B$, therefore $A\setminus B=\emptyset \implies \vol(A\setminus B)=0$, leading to a contradiction to our initial claim that $\vol(A\setminus B)>0$. Therefore our initial claim must be wrong, and $\vol(A\setminus B)=0$. Using the same argument $\vol(B\setminus A)=0$, implying that $A=B$ almost everywhere.
\end{proof}

\begin{proof}[\textbf{Proof of Theorem~\ref{thm:deter-prob-FRS}}]
    We establish this proof by showing that any measurable subset of $F_t$ with non-zero measure is also a subset of $\omega_t^*$, and every measurable subset of $\omega_t^*$ with non-zero measure is a subset of $F_t$ and then invoking Lemma~\ref{lem:subset}.

    Let $\omega\subseteq\omega_t^*$ be a measurable set such that $\vol(\omega)>0$. It follows that $\mu_t(\omega)>0$ for if it doesn't, then $\mu_t(\omega_t^*\setminus\omega)=1$, but $\vol(\omega_t^*\setminus\omega) < \vol(\omega_t^*)$ leading to a contradiction with the definition of $\omega_t^*$ in \eqref{eq:probabilistic-FRS}. Since $\mu_t(\omega)>0$, by the properties of push-forward measure, there exist subsets $\mathcal{U}_i$ and $\mathcal{W}_i$ with non-zero measure that can bring the agent's state to $\omega$. Therefore, $\omega \subseteq F_t$.

    Let $F\subseteq F_t$ be a measurable set with $\vol(F)>0$. 
    As the composition of $f$ in \eqref{eq:deterministic-FRS} for a fixed $x_0$---for the sake of convenience, let's call it $\hat{f}$---is a continuously differentiable map, by the Luzin's N-property for multi-dimensional maps, the pre-image $\hat{f}^{-1}(F)\subseteq\mathcal{U}_{t-1}\times\cdots\mathcal{U}_0\times\mathcal{W}_{t-1}\times\cdots\times\mathcal{W}_0$ will be a set of non-zero Lebesgue measure as well. As we assume that the distribution over the space of controls and disturbances is absolutely continuous, and $\hat{f}^{-1}(F)$ lies in the support of these distributions with non-zero Lebesgue measure, it follows that the probability measure of $\hat{f}^{-1}(F)$ is also non-zero. By the property of the push-forward measure if the probability measure of $\hat{f}^{-1}(F)$ is non-zero, then the measure under the push-forward of these sets is also non-zero, i.e., $\mu_t(F)>0$. Hence, $F\subseteq\omega_t^*$, for if it were removed from the set, the feasibility constraint of the lower bound will be violated.
\end{proof}

\begin{proof}[\textbf{Proof of Theorem~\ref{thm:convex-opt}}]
    The volume of an ellipse $E_i(c_i)$ in an $n$-dimensional space \cite{wilson2010volume} is given by 
    \begin{equation}
        \vol(E_i(c_i)) = \frac{\pi^{n/2}c_i^{n/2}}{\Gamma(n/2+1)}\sqrt{\lambda_{i,1}\cdots\lambda_{i,n}}
    \end{equation}
    where $\Gamma$ is the standard gamma function in calculus while $\lambda_{i,j}$ are the eigen values of $\Sigma$. Setting $n=2$ in the above and using it in \eqref{eq:approx-opt} gives the objective function of \eqref{eq:convex-opt}.

    In an $n$-dimensional space the function $V_i(x)$ is $\chi^2$ distributed if $x$ is drawn from a Gaussian \cite{zorin1989distribution}. For the special case of $n=2$, the cumulative distribution of $\chi^2$ gives $\mu_{t,i}(E_i(c_i)) = (1-\exp(-c_i/2))$. Using this in \eqref{eq:approx-opt} gives the constraint of \eqref{eq:convex-opt}.

    Finally, we note that the objective function of \eqref{eq:convex-opt} is linear, while the constraint is convex. For the latter assertion, note that the second derivative of $p_i (1-\exp(-c_i/2))$ is $-p_i \exp(-c_i/2)/4$ which is always negative, hence $p_i (1-\exp(-c_i/2))$ is concave for all $i$. Sum of concave functions is concave, therefore $\sum_{i=1}^K p_i (1-\exp(-c_i/2))$ is concave. Since the super-level sets of concave functions are convex, we get that the constraint of \eqref{eq:convex-opt} is also convex, implying that \eqref{eq:convex-opt} is a convex optimization problem.
\end{proof}

\begin{proof}[\textbf{Proof of Corollary~\ref{cor:scale-invariance}}]
    We will establish this corollary by contradiction. Let $\alpha_1,\alpha_2\in(0,\infty)$ and $\alpha_1\neq \alpha_2$. Let $\{c_{i,\alpha_1}^*\}_{i=1}^K$ and $\{c_{i,\alpha_2}^*\}_{i=1}^K$ be the solutions of \eqref{eq:convex-opt} for scaled covariances $\{\alpha_1\Sigma_i\}_{i=1}^K$ and $\{\alpha_2\Sigma_i\}_{i=1}^K$, respectively. Let's assume that there exists an $i$, such that $c_{i,\alpha_1}^*\neq c_{i,\alpha_2}^*$, and without loss of generality assume that $c_{i,\alpha_1}^*< c_{i,\alpha_2}^*$. Then, 
    \begin{align}\nonumber
        \sum_{i=1}^K \pi \alpha_2 \sqrt{\lambda_{i,1}\lambda_{i,2}} c_{i,\alpha_1}^* < \sum_{i=1}^K \pi \alpha_2 \sqrt{\lambda_{i,1}\lambda_{i,2}} c_{i,\alpha_2}^*.
    \end{align}
    Further, observe that $\sum_{i=1}^K p_i (1-\exp(-c_{i,\alpha_1}^*/2))\geq \tau$ as it is an optimal (hence, feasible) solution of \eqref{eq:convex-opt} for $\{\alpha_1\Sigma_i\}_{i=1}^K$, therefore, $c_{1,\alpha_1}^*,\cdots,c_{K,\alpha_1}^*$ is also a feasible solution of \eqref{eq:convex-opt} for $\{\alpha_1\Sigma_i\}_{i=1}^K$, but attains a smaller objective function value than the optimal solution of \eqref{eq:convex-opt} for $\alpha_2$, leading to a contradiction, hence, the claim must be true.
\end{proof}

\begin{proof}[\textbf{Proof of Lemma~\ref{lem:monotonic-set-inclusion}}]
    Using Corollary~\ref{cor:scale-invariance}, it follows that $\{c_i^*\}_{i=1}^K$ are the same regardless of the scaling choice. As $\alpha_1 < \alpha_2$, then it follows that $\{x~|~(x-\overline{x}_i)^{\rm T} \alpha_1^{-1}\Sigma_i^{-1} (x-\overline{x}_i) \leq c_i^*\} \subseteq \{x~|~(x-\overline{x}_i)^{\rm T} \alpha_2^{-1}\Sigma_i^{-1} (x-\overline{x}_i) \leq c_i^*\}$ as $(x-\overline{x}_i)^{\rm T} \alpha_1^{-1}\Sigma_i^{-1} (x-\overline{x}_i) > (x-\overline{x}_i)^{\rm T} \alpha_2^{-1}\Sigma_i^{-1} (x-\overline{x}_i)$. Therefore, taking the union of all these sets gives us $E^*(\{\alpha_1\Sigma_i\}_{i=1}^K) \subseteq E^*(\{\alpha_2\Sigma_i\}_{i=1}^K)$.
\end{proof}

\end{document}